\pdfoutput=1

\documentclass[11pt]{article}

\usepackage[dvipsnames]{xcolor}
\usepackage[]{EMNLP2022}
\usepackage{booktabs}
\usepackage{CJK}
\usepackage{amsmath}
\usepackage{enumitem}
\usepackage{amsfonts}
\usepackage{multirow}
\usepackage{times}
\usepackage{latexsym}
\usepackage{natbib}
\usepackage{array}
\usepackage{tabularx}
\newcommand*{\rom}[1]{\expandafter\@slowromancap\romannumeral #1@}

\usepackage[T1]{fontenc}

\usepackage[utf8]{inputenc}
\usepackage{graphicx}
\usepackage{subfigure}
\usepackage{float}
\usepackage{mathtools}
\usepackage{bm}
\usepackage{amssymb}
\usepackage{pifont}
\usepackage{microtype}
\usepackage{color}

\title{Getting the Most out of Simile Recognition}

\author{Xiaoyue Wang\textsuperscript{1, 2\footnotemark[1]}~~~Linfeng Song\textsuperscript{3\footnotemark[1]}~~~Xin Liu\textsuperscript{1}~~~Chulun Zhou\textsuperscript{1}~~~Jinsong Su\textsuperscript{1, 2\footnotemark[2]}\\
  \textsuperscript{1}School of Informatics, Xiamen University, China \\
  \textsuperscript{2}Key Laboratory of Digital Protection and Intelligent Processing of Intangible \\ Cultural Heritage of Fujian and Taiwan (Xiamen University), Ministry of Culture and Tourism, China\\
  \textsuperscript{3}Tencent AI Lab, Bellevue, WA, USA\\
  
  \texttt{\small{xiaoyuewang@stu.xmu.edu.cn}}~~~
  \texttt{\small{lfsong@tencent.com}}\\
  \texttt{\small{\{liuxin, clzhou\}@stu.xmu.edu.cn}}~~~
  \texttt{\small{jssu@xmu.edu.cn}}
  }

\begin{document}
\maketitle
\begin{abstract}
Simile recognition involves two subtasks: \emph{simile sentence classification} that discriminates whether a sentence contains simile, and \emph{simile component extraction} that locates the corresponding objects (i.e., tenors and vehicles).
Recent work ignores features other than surface strings.
In this paper,
we explore expressive features for this task to achieve more effective data utilization.
Particularly, we study two types of features: 1) input-side features that include POS tags, dependency trees and word definitions, and 2) decoding features that capture the interdependence among various decoding decisions.
We further construct a model named \emph{HGSR}, which merges the input-side features as a heterogeneous graph and leverages decoding features via distillation.
Experiments show that \emph{HGSR} significantly outperforms the current state-of-the-art systems and carefully designed baselines, verifying the effectiveness of introduced features. Our code is available at \url{https://github.com/DeepLearnXMU/HGSR}.

\end{abstract}

\renewcommand{\thefootnote}{\fnsymbol{footnote}}
\footnotetext[1]{Equal contribution.}
\footnotetext[2]{Corresponding author.}

\section{Introduction}
\label{intro}
Simile is a type of figurative that compares two objects (named \emph{tenor} and \emph{vehicle}) of different categories using comparator words such as ``\emph{like}'', ``\emph{as}'' or ``\emph{than}''.
Table \ref{data_exp} shows a simile sentence, where the tenor ``\emph{sheep}'' and the vehicle ``\emph{clouds}'' are compared using comparator ``\emph{like}''.
Generally, simile recognition involves two subtasks \cite{liu2018neural}: \emph{simile sentence classification} that discriminates whether a sentence contains simile expressions, and \emph{simile component extraction}, which aims to find simile components (i.e., tenor and vehicle).
Because simile usually involves implicit sentiment, 
it can provide essential information for sentiment analysis \cite{sentiment1,sentiment2} (e.g. hate speech detection) and dialogue understanding \cite{dialogue2019}.
Besides, simile recognition can help language learners to better understand the implicit meanings expressed by the simile expressions in novels and fairy tale stories. Therefore, simile recognition has become an important task in natural language processing.
\begin{table}[t]
\centering
\begin{CJK}{UTF8}{gbsn}
\begin{tabular}{l}
\toprule[1pt]
\begin{tabular}[l]{@{}l@{}}\textcolor{RoyalBlue}{羊群}看上去\textbf{像}白\textcolor{PineGreen}{云}。\\ (\emph{The \textcolor{RoyalBlue}{sheep} look like white \textcolor{PineGreen}{clouds}.})\end{tabular} \\ \midrule
\begin{tabular}[l]{@{}l@{}}她看上去\textbf{像}我姐姐。\\ (\emph{She looks like my sister.})\end{tabular}\\ \bottomrule[1pt]
\end{tabular}
\end{CJK}
\caption{Two examples: the first is a simile sentence that uses ``\emph{like}'' to compare tenor ``\emph{\textcolor{RoyalBlue}{sheep}}'' and vehicle ``\emph{\textcolor{PineGreen}{clouds}}''; the second is a literal sentence.}
\label{data_exp}
\vspace{-1.0em}
\end{table}


Previous studies on simile recognition \cite{niculae2013comparison,niculae2013computational,niculae2014brighter} demonstrate that exploiting syntactic features is beneficial to simile recognition.
However, they resort to handcrafting feature templates, which usually requires extensive efforts from linguistic experts and is hard to be adapted to new domains and languages.
With the recent release of annotated data in a descent scale and the success of deep learning on a wide range of NLP tasks, \citet{liu2018neural} first propose a neural model for simile recognition.
Specifically, they adopt multi-task learning and let the two subtasks share an LSTM \cite{hochreiter1997long} encoder that only consumes input sentences.
Along this line,
\citet{zeng2020neural} propose a cyclic multitask learning model with a pretrained BERT \cite{devlin2019bert} encoder, where both subtasks and an extra language modeling subtask are stacked into a loop.
This cyclic model yields the current state-of-the-art (SOTA) performance.
In spite of these successful attempts, they suffer from the data hunger issue, and ignore other features except surface strings.

\begin{figure*}[t]
\centering
    \subfigure[]{
    \includegraphics[width=0.60\linewidth]{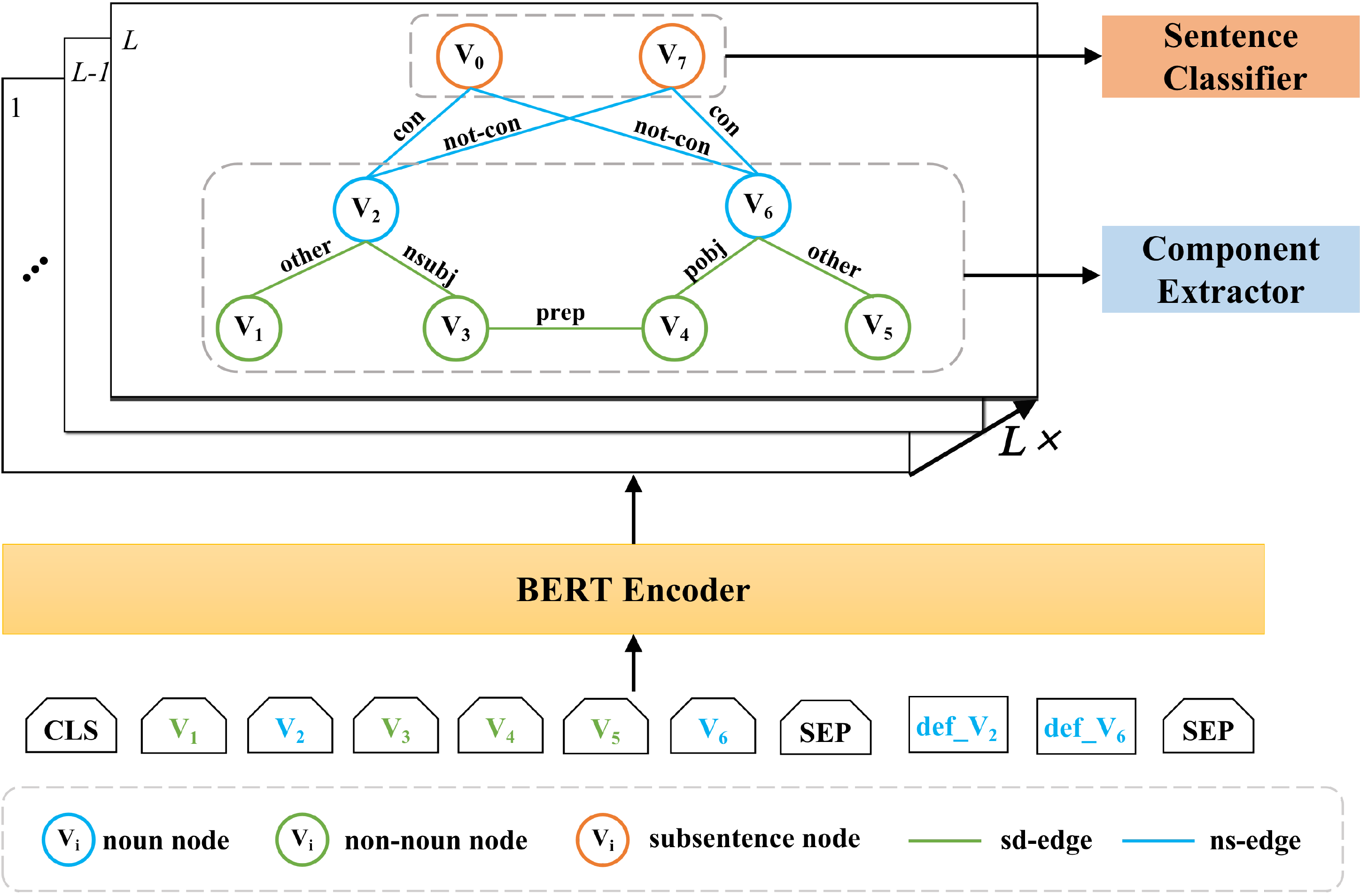}
        \label{model}
    }
    \subfigure[]{
    \includegraphics[width=.35\linewidth]{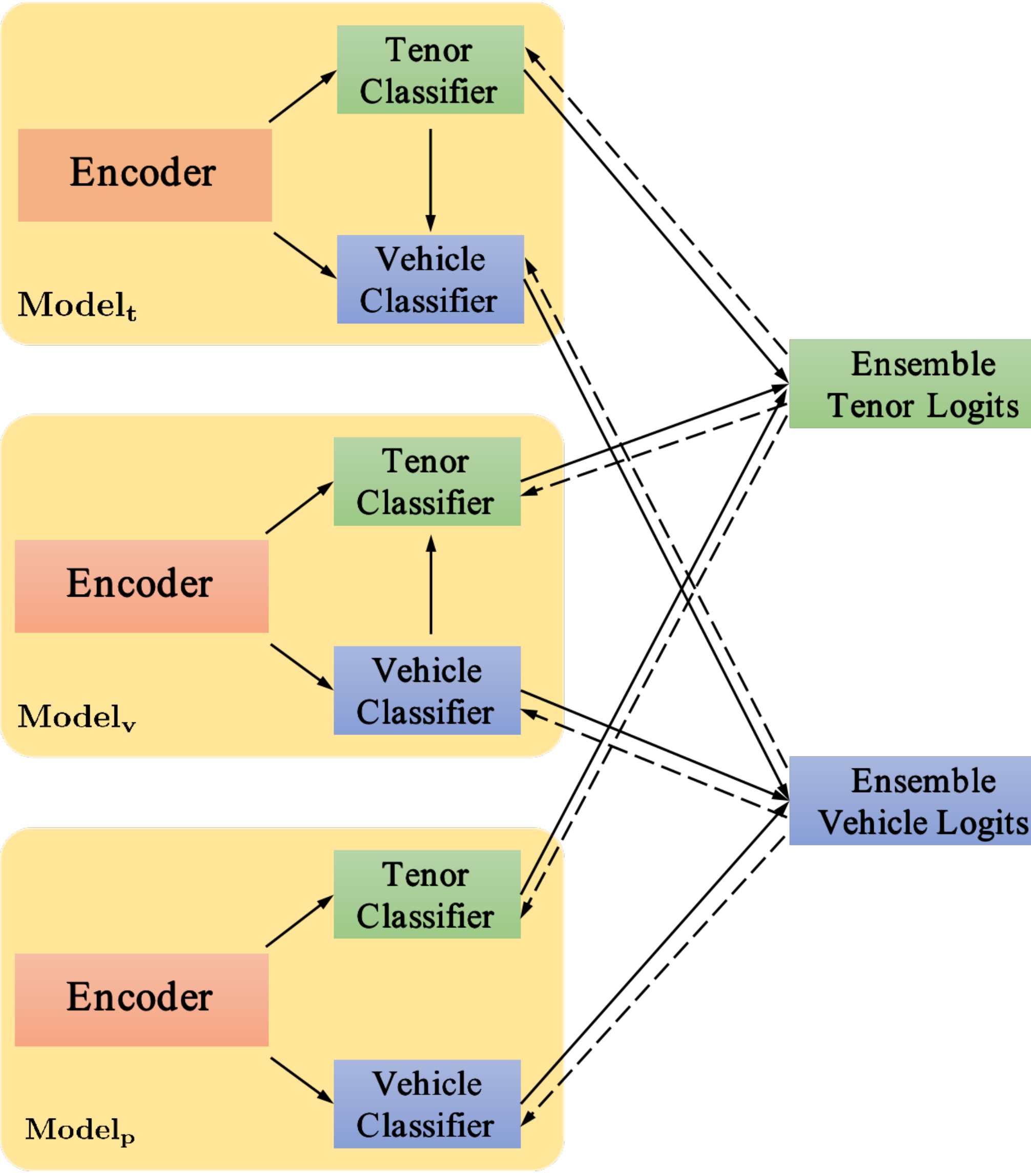}
        \label{model_b}
    }
    \caption{(a) The architecture of our model taking a heterogeneous graph constructed from input-side features, and (b) The process of distillation using decoding features. The decoding orders of the three models are ``tenor $\rightarrow$ vehicle'', ``vehicle $\rightarrow$ tenor'' and ``in parallel'', respectively.}
\end{figure*}

In this paper, we explore more expressive features to achieve more effective data utilization for neural simile recognition.
The studied features can be categorized into two major types: one type (\textbf{input-side features}) covers the task input, and the other type (\textbf{decoding features}) captures the interdependence among various decoding decisions.
Particularly, our input-side features include POS tags, dependency trees and word definitions,
and we propose a novel heterogeneous graph that used
to effectively merge the input-side features.
In the heterogeneous graph, some nodes represent input words, and we use POS tags to distinguish \textit{noun} nodes (in blue) from \textit{non-noun} nodes (in green) as shown in \ref{model}.
The noun words are highlighted because simile components are usually nouns \cite{hanks2012roles}, and their dictionary definitions (if any) are added to help learn their representations.
We also introduce two \textit{subsentence} nodes divided by the given comparator (e.g., ``\emph{like}'') to help contrast the both sides of the comparator.
Meanwhile, each edge may correspond to a dependency arc (e.g., ``nsubj'') or point from a noun node to a subsentence node.
We use multiple GAT \cite{gat2017graph} layers to represent each graph.



We introduce the decoding features for simile component extraction. As the tenor and the vehicle are different entities with the same properties \cite{niculae2013comparison}, intuitively, the tenor information can help to locate vehicle, and vice versa. 
To model such intuition,
we sequentially extract tenor and vehicle, where the encoder states of the first extracted component (e.g., the tenor) are consumed as extra decoding features for recognizing the second component (e.g., the vehicle).
To leverage all possible decoding features, we take the ensemble of the models for all three decoding orders (tenor $\rightarrow$ vehicle; vehicle $\rightarrow$ tenor; in parallel) as the teacher model.
The teacher model then simultaneously guide each individual model via distillation during training.
During inference time, we only use one model to avoid the computational consumption caused by their ensemble.

Extensive experiments on a simile recognition benchmark \cite{liu2018neural} show that our proposed model largely outperforms previous SOTA system and several competitive baselines by 1.7 and 9.3 points for simile sentence classification and simile component extraction, respectively.
Besides, our model trained with 40\% data reaches comparable performances than the baseline using full training data, indicating that our model is less data hungry.

\section{Problem Formulation}
\label{sec:def}
Formally, given an input sentence $S = w_{1}, \dots, w_{i},$ $\dots, w_{N}$ containing a comparator $w_c$, the goal is to detect whether the comparator $w_c$ is a simile and what spans in $S$ correspond to the simile components (tensors and vehicles).
Note that a comparator may correspond to a literal comparison (rather than a simile), such as ``\emph{the sheep looks like an Australian sheep breed}.''
Following previous work \cite{niculae2014brighter,liu2018neural,simile2019,zeng2020neural}, we formulate the two subtasks as binary classification and sequence labeling, respectively.

\section{Baseline: BERT for Simile Recognition}
\label{sec:bertmodel}

In this section, we introduce a baseline for simile recognition based on BERT \cite{devlin2019bert}, temred as \textbf{BSR}.



\subsection{Encoding}
\label{sec:baseline_encoding}
Given an input sentence $S = w_{1}, \dots, w_{i}, \dots, w_{N}$, we first place special tokens \texttt{[CLS]} and \texttt{[SEP]} at its beginning and the ending, before feeding the sequence into a BERT encoder with extra self-attention \cite{vaswani2017attention} layers.
Let $\textbf{H}=\left\{h_{i}\right\}_{i=0}^{N+1}$ be the hidden states of tokens at the top layer, the hidden state ($h_{0}$) of \texttt{[CLS]} is used as the sentence representation.

\subsection{Simile Sentence Classification}
\label{sec:baseline_classification}
For simile classification, we feed the sentence representation $h_{0}$ into a linear layer:
\begin{equation}
 \label{eq:baseline_simile_class}
p(c|S) = \operatorname{softmax}(W_{c} h_{0}),
\end{equation}
where $c\in\{\mathtt{true}, \mathtt{false}\}$ is the label indicating whether the input sentence $S$ contains simile and $W_{c}$ is a model parameter.
The corresponding loss is defined as:
\begin{equation}
    \mathcal{J}_{s c}(c|S;\theta) = -\log p(c|S)
\end{equation}

\subsection{Simile Component Extraction}
\label{sec:baseline_extraction}
Afterwards, we feed the final node states $\textbf{H}$ into a fully-connected layer with $\operatorname{softmax}$ to conduct component extraction: 
\begin{equation}
\begin{aligned}
\label{eq5}
p(T|S) &= \prod_{i=1}^N p(t_i|S) \\ 
&= \prod_{i=1}^N \operatorname{softmax}(W_{e}\cdot h_i + b_e), \\
\end{aligned}
\end{equation}
where $T=t_1,\dots,t_N$ is the gold label sequence of simile component extraction. The possible values for each $t_i$ are $\{\mathtt{T}, \mathtt{V}, \mathtt{O}\}$, indicating $w_i$ being part of a tenor, part of a vehicle and none, respectively.
The loss is defined as:
\begin{equation}
    \mathcal{J}_{c e}(T|S;\theta) = -\sum_{i=1}^{N}\log p(t_i|S),
\end{equation}

\subsection{Training}
Given training data $\mathcal{D}$, we train the model by a linear interpolation between the two subtasks:

\begin{equation}
\begin{split}
\label{eq:baseline_objective}
    \mathcal{J}(\mathcal{D};\theta) = \sum_{(S,c,T) \in \mathcal{D}} \Big( \alpha \cdot \mathcal{J}_{s c}(c|S;\theta)+\\
    (1 - \alpha) \cdot \mathcal{J}_{c e}(T|S;\theta) \Big),
\end{split}
\end{equation}
where $\theta$ denotes all model parameters, and $\alpha$ is the interpolation coefficient between the two subtasks.

\section{Model}
\label{section3}
In this section, we give more details of our model that takes input-side features (\S \ref{sec:ifeatures}) and decoding feature (\S \ref{sec:dfeatures}) for simile recognition.
For fair comparison, our model is mostly consistent with the baseline (\S \ref{sec:bertmodel}) with slight changes (shown below) for incorporating our introduced features.

\subsection{Including Input-side Features}
\label{sec:ifeatures}
We explore the following three types of input-side features to enhance each input sentence:
\begin{itemize}[leftmargin=*]
\item \textbf{POS Tags}: We mainly use the part-of-speech (POS) information to distinguish nouns and other words in each given sentence. This is because simile components are usually nouns.
\item \textbf{Dependency Tree}:
Dependency trees have been shown to capture long-range word-to-word dependencies and some shallow semantic information. We adopt such knowledge to help our model learn better sentence representations.

\item \textbf{Word Definitions}:
We adopt a word sense analyzer \cite{DBLP:conf/emnlp/YaoPJCYY21} to find definitions for the nouns.
The definitions are then appended as input features to our model.
Intuitively, using definitions can improve model robustness and achieve more effective data utilization.
It also shares a similar spirit with recent prompt-based research \cite{DBLP:conf/eacl/SchickS21}.
\end{itemize}

\paragraph{Combination with Heterogeneous Graph}
We merge the input-side features for each instance by constructing a heterogeneous graph $G=(V, E)$, which includes the set of nodes $V$ and edges $E$:

\subparagraph{Node.}
\label{nodes}In the node set $V$, each node may represent a noun, a non-noun word or a subsentence. This is based on the POS tagging results by a pretrained LTP parser\footnote{https://github.com/HIT-SCIR/ltp}.
As simile usually involves the comparison between a tenor and a vehicle located on both sides of a comparator, we introduce two subsentence nodes that correspond to the left and right parts split by the comparator, respectively.
Taking Figure \ref{model} as the example, the graph contains two noun nodes (\textcolor{cyan}{$v_2$} and \textcolor{cyan}{$v_6$}), four non-noun nodes (\textcolor{ForestGreen}{$v_1$}, \textcolor{ForestGreen}{$v_3$} \textcolor{ForestGreen}{$v_4$}, and \textcolor{ForestGreen}{$v_5$}) and two subsentence nodes (\textcolor{orange}{$v_0$} and \textcolor{orange}{$v_7$}).

The reason for the above design is to better determine whether the subsentences divided by the comparator contain different types of objects with similar attributes, which is crucial for simile sentence classification. We expect that the subsentence nodes can help highlight the difference between the two sides of a comparator during information aggregation within the graph neural network.
Moreover, since the considered objects are usually nouns, we believe that nouns are more important than other words in this task. Thus, we deliberately differentiate noun and non-noun nodes to emphasize the positive impact of nouns.

\subparagraph{Edges.}
We consider two main types of edges in the edge set $E$.
The edges of the first type (named \emph{sd-edge}) are essentially dependency arcs.
To avoid excessive trainable parameters, we restrict the edges to only cover the top 8 most frequent dependency relations in the training data.
Meanwhile, we convert all rest dependency relations as ``\emph{other}''.
By doing so, we expect that the \emph{sd-edges} are able to capture the long-distance dependency between each tenor and its vehicle.
An edge of the second type (named \emph{ns-edge} that is short for \emph{noun-subsentence edge}) connects a noun node with a subsentence node.
In this way, the impacts of nouns are highlighted when aggregating information for subsentence nodes. 
In order to distinguish the two subsentence nodes, we assign each \emph{ns-edge} with a label that can either be ``\emph{con}'' or ``\emph{not-con}'', indicating whether the subsentence contains the noun.
Using Figure \ref{model} as the example,
the labels of many \emph{sd-edges} are set to \emph{``other''} except for ``\emph{nsubj}'', ``\emph{prep}'' and ``\emph{pobj}''.
For the \emph{ns-edges}, as ``$\textcolor{cyan}{v_2}$'' belongs to the left subsentence, the edge labels for ``$\textcolor{cyan}{v_2} \rightarrow \textcolor{orange}{v_0}$'' and ``$\textcolor{cyan}{v_2} \rightarrow \textcolor{orange}{v_7}$'' are \emph{``con''} and \emph{``not-con''}, respectively.

\paragraph{Heterogeneous Graph Encoding}
As shown in Figure \ref{model}, we modify the baseline encoding phase (\S \ref{sec:baseline_encoding}) by replacing the extra self-attention layers with GAT \cite{gat2017graph} layers in order to consume the proposed heterogeneous graphs.
The initial state (e.g., $g_i^{(0)}$) for a word node (in \textcolor{cyan}{blue} and \textcolor{ForestGreen}{green}) is initialized from the corresponding BERT output ($h_i$).
For a subsentence node (in \textcolor{orange}{orange}), we initial its state using average pooling over the hidden states of the words within the subsentence.
The embeddings (e.g, $e_{i j}$) for the edge labels are randomly initialized.



At each GAT layer, we sequentially conduct graph attention and gating mechanisms to update all node states. 
Taking the $l$-th layer for example, we first update each $g_{i}^{(l)}$ with the hidden states (e.g., $g_{j}^{(l)}$) of its directly connected neighbors as follows: 
\begin{equation}
    z_{i j}=\operatorname{LeakyReLU}(W_{a}[W_{q} g_{i}^{(l)};W_{k} g_{j}^{(l)};e_{i j}]), \notag
\end{equation}
\vspace{-1.0em} 
\begin{equation}
\begin{aligned}
    & \alpha_{i j}=\operatorname{softmax}(z_{i j})=\frac{\exp(z_{i j})}{\sum_{k \in \mathcal{N}_{i}}\exp(z_{i k})}, \\
    & g_{i}^{(l+1)}=\sigma\left(\sum_{j \in \mathcal{N}_{i}} \alpha_{i j} W_{v} g_{j}^{(l)}\right),
\end{aligned}
\end{equation}
where $\alpha_{i j}$ is the attention score indicating the importance of node $j$ to node $i$, $\mathcal{N}_{i}$ is the set of neighborhood nodes to the node $i$ in the graph, $W_{*}$ are model parameters\footnote{For the remaining of this paper, we use $W_{*}$ and $b_{*}$ to denote model parameters.}, and $\sigma(*)$ is a sigmoid function.

Note that different from the BERT-based simile recognition model, which has only one sentence representation (\S  \ref{sec:baseline_classification}), there are two subsentence representations ($g_{0}^{(L)}$ and $g_{N+1}^{(L)}$). Hence, we adjust Eq. \ref{eq:baseline_simile_class} for simile sentence classification as follows:
\begin{multline} \label{eq:simile_class}
p(c|S) = \operatorname{softmax}(W_{c}[g_{0}^{(L)};g_{N+1}^{(L)};\\
|g_{0}^{(L)} - g_{N+1}^{(L)}|]E_{c}^\top),
\end{multline}

\subsection{Leveraging Decoding Features}
\label{sec:dfeatures}
As shown in Figure \ref{model_b}, we adopt two extra models to extract simile components (tenor and vehicle) sequentially: one extracts the \emph{tenor} before the \emph{vehicle} (model$_t$) while the other functions in the opposite direction (model$_v$).
Different from the baseline, which extracts simile components in parallel (model$_p$), both models use the encoder state of the first component as extra features to the second component extractor.
In particular, the extractor for the second component is defined as
\begin{equation}
\begin{aligned}
\label{eq:second_component}
p(T_{c_2}|S) &= \prod_{i=1}^N p(t_{c_2,i}|S) \\ 
&= \prod_{i=1}^N \operatorname{softmax}(W_{e}^{c_2}\cdot [g_i^{(L)}; g_{c_1}^{(L)}] + b_e^{c_2}),
\end{aligned}
\end{equation}
where $T_{c_2}$ is the gold label sequence for extracting the second simile component, and $g_{c_1}^{(L)}$ denotes the hidden state for the first simile component. Both $W_{e}^{c_2}$ and $b_e^{c_2}$ represent the parameters for the extra simile component extractor.


To leverage all possible features, inspired by \cite{DBLP:journals/taslp/ZhangXSL19,DBLP:conf/aaai/WuCGLZS22}, we apply distillation to encourage each model to mimic the behaviors of their
ensemble: 
\begin{equation}
\begin{aligned}
    \mathcal{J}_{kl}^*(\mathcal{D};\theta^*) =& \operatorname{KL}(p_*(T|S)||p_{ensemble}(T|S))
\end{aligned}
\end{equation}
where $p_*(T|S)$ and $\theta^*$ are the output probabilities and parameters for one individual model, $\operatorname{KL}(*)$ denotes the Kullback-Leibler divergence, and
$\mathcal{J}_{kl}^*(\mathcal{D};\theta^*)$ is the distillation objective.
We sum up the logits of all models as their ensemble, and $p_{ensemble}(T|S)$ denotes the probability for the ensemble.
The final objective of each model can be denoted as:
\begin{equation}
\begin{aligned}
    \mathcal{J}_{final}^*(\mathcal{D};\theta^*) &= \lambda \mathcal{J}^*(\mathcal{D};\theta^*) +  (1-\lambda)\mathcal{J}_{kl}^*(\mathcal{D};\theta^*) \\
    &=\lambda \sum_{(S,c,T) \in \mathcal{D}} \Big( \alpha \cdot \mathcal{J}_{s c}(c|S;\theta)+\\
    &(1 - \alpha) \cdot \mathcal{J}_{c e}(T|S;\theta) \Big) + (1-\lambda) \\ &\operatorname{KL}(p_*(T|S)||p_{ensemble}(T|S))
\end{aligned}
\end{equation}
where $\mathcal{J}^*(\mathcal{D};\theta^*)$ is the training objective defined in Eq.\ref{eq:baseline_objective}, and $\lambda$ is the hyper-parameter to control the impacts of two objectives. Here we linearly increase $\lambda$ from 0 to 1 throughout training. It is worth noting that for less consumption, only one model is used during inference time.

\section{Experiments}

We conduct detailed experiments and analysis to investigate the effectiveness of our model.

\subsection{Setup}

\subparagraph{Datasets.} We choose the Chinese Simile Recognition benchmark \cite{liu2018neural}, which consists of 11,377 sentences (roughly half of them contain simile).
Since no data split on training, developing and testing is provided, we follow previous work to conduct 5-fold cross validation\footnote{In our experiments, the standard deviations of the 5-fold cross validation for simile sentence classification and simile component extraction are 0.29 and 0.32, respectively.}. 
We also follow previous work to evaluate our models using the official scorer that measures Precision, Recall and F1 score.


\subparagraph{Comparisons.}
To comprehensively evaluate the \textbf{BSR} baseline and our \textbf{HGSR} model, we compare them with the following systems:
\begin{itemize}[leftmargin=*]
\setlength{\itemsep}{0pt}
\item{\textbf{MTL} \cite{liu2018neural}. 
    A multi-task learning model, where simile sentence classification, simile component extraction and sentence reconstruction are jointly modeled.}
\item{\textbf{Self\_Attn+POS} \cite{simile2019}. 
    It extends \cite{liu2018neural} with POS information and uses several self-attention layers to enhance the original LSTM encoder. 
    } 
\item{\textbf{Cyc-MTL} \cite{zeng2020neural}. It extends \cite{liu2018neural} by stacking the three subtasks into a cycle to let them better benefit from each other. 
}
\end{itemize}

To verify that our heterogeneous graph can effectively incorporate POS features, we also build a variant of our  model (HGSR-ConcatPOS), which removes noun nodes from the graph and concatenates each word embedding with its POS embedding.


\begin{table*}[t]
\centering
\begin{tabular}{l|ccc|ccc}
\toprule[1pt]
\multirow{2}{*}{Model} & \multicolumn{3}{c|}{Sentence Classification}               & \multicolumn{3}{c}{Component Extraction}                   \\ \cline{2-7} 
                       & \multicolumn{1}{c}{Precision} & \multicolumn{1}{c}{Recall} & F1 & \multicolumn{1}{c}{Precision} & \multicolumn{1}{c}{Recall} & F1 \\ \hline

MTL \cite{liu2018neural}              & 80.84 & 92.20 & 86.15 & 61.60 & 73.61 & 67.07 \\
Self\_Attn+POS \cite{simile2019} & 80.44 & 91.69 & 85.70 & 58.91 & 74.65 & 65.85 \\
Cyc-MTL \cite{zeng2020neural}          & 85.81 & \textbf{94.43} & 89.92 & 73.97 & 77.61 & 75.74 \\
\hline
HGSR-ConcatPOS              &  88.73     &   94.30    &    91.43   &  81.23     &  87.76     & 84.37   \\ 

HGSR              &  \textbf{89.04}     &   94.39    &    \textbf{91.64}   &  \textbf{81.86}     &  \textbf{88.37}     & \textbf{84.99}   \\ \bottomrule[1pt]
\end{tabular}%
\caption{Main test results. Please note that we outperform all baselines, including the SOTA Cyc-MTL.}
\label{table1}
\vspace{-0.5em}
\end{table*}



\subparagraph{Implementation Details.}
We determine hyperparameters $\alpha$ as 0.1 according to the model performance on the validation set. Following \citet{zeng2020neural}, we employ a pre-trained Chinese BERT\footnote{https://github.com/ymcui/Chinese-BERT-wwm} model to learn contextual word embeddings and then finetune this model using our training data. Besides, we randomly initialize the representation vectors for edges and label embeddings with 50 and 100 dimension vectors, respectively. The max length and word number of sentence are set to be 128 and 100 by padding shorter sentences and cutting longer ones. The hidden state size of the GAT layer is set to 300. Parameter optimization is performed using Adam with learning rate 2e-5 and batch size 8. And we stack two layers of GAT to gather global information after conducting simile sentence classification. For fair comparison, we also stack two self-attention layers on the \emph{BSR} baseline.
After training, we evaluate model$_t$, model$_v$ and model$_p$ on the development set and pick up the best one for inference.

\subsection{Effect of GAT Layer Number $L$}

We first investigate the effect of the GAT layer number $L$ on the development set. When $L$ varies from 1 to 3, the F1 scores of our model on simile classification are 91.32, 92.44, and 92.32, those on component extraction are 86.74, 87.53, and 87.49, respectively. Therefore, we set $L$ as 2 in subsequent experiments.

\subsection{Main Results}

The main test results are shown in Table \ref{table1}. We can observe \emph{HGSR} outperforms all the baselines and achieves SOTA $F_{1}$ scores on both subtasks, which demonstrates the effectiveness of our methods. In order to further understand advantages of \emph{HGSR}, we follow \citet{zeng2020neural} to conduct more evaluations. 


\begin{figure}[!t]
 \includegraphics[width=\linewidth]{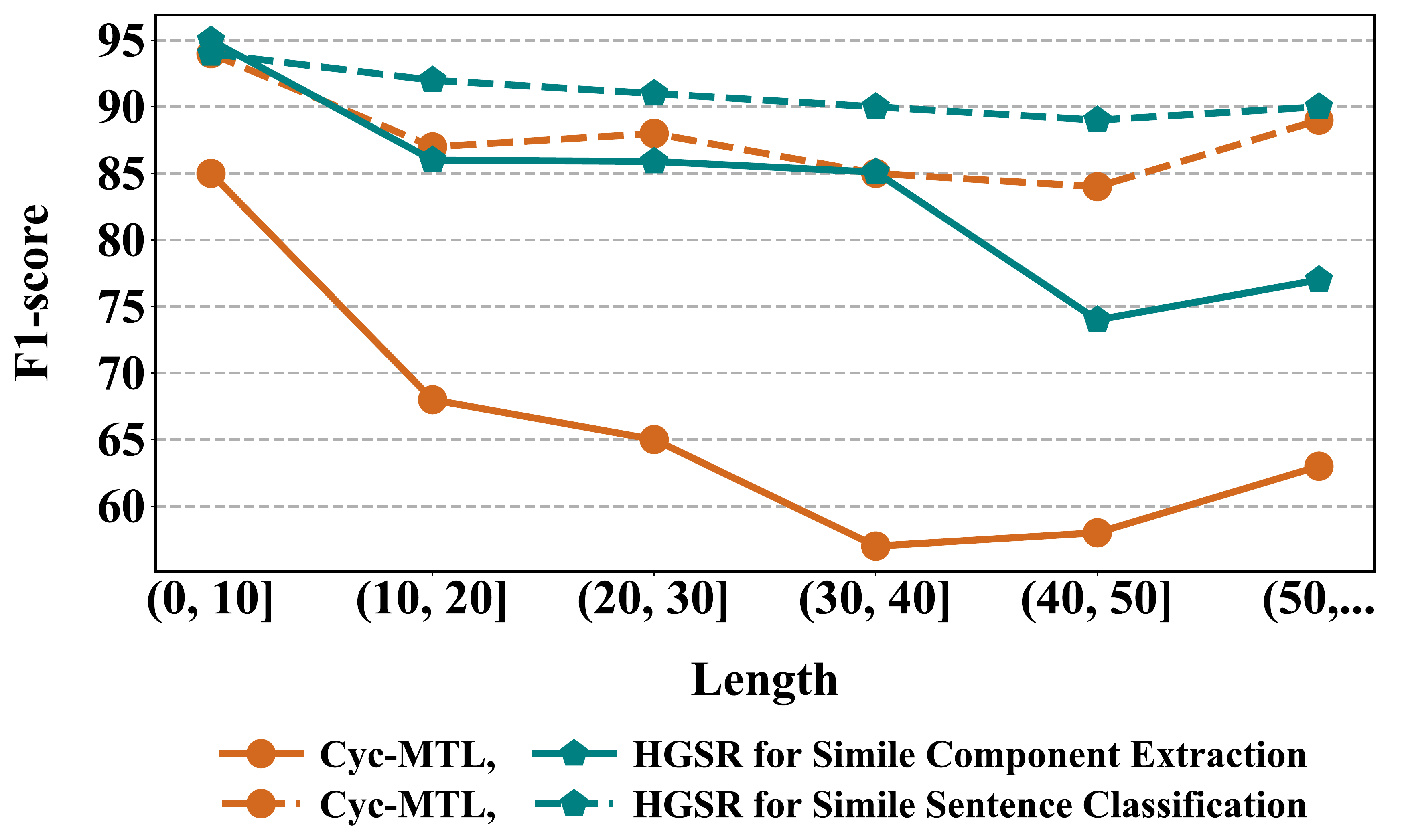}
 \caption{F1 scores on different groups of test instances according to sentence lengths. Dashed lines and solid lines represent simile sentence classification and simile component extraction, respectively.}
 \label{length}
\end{figure}

\begin{figure}[t]
 \includegraphics[width=\linewidth]{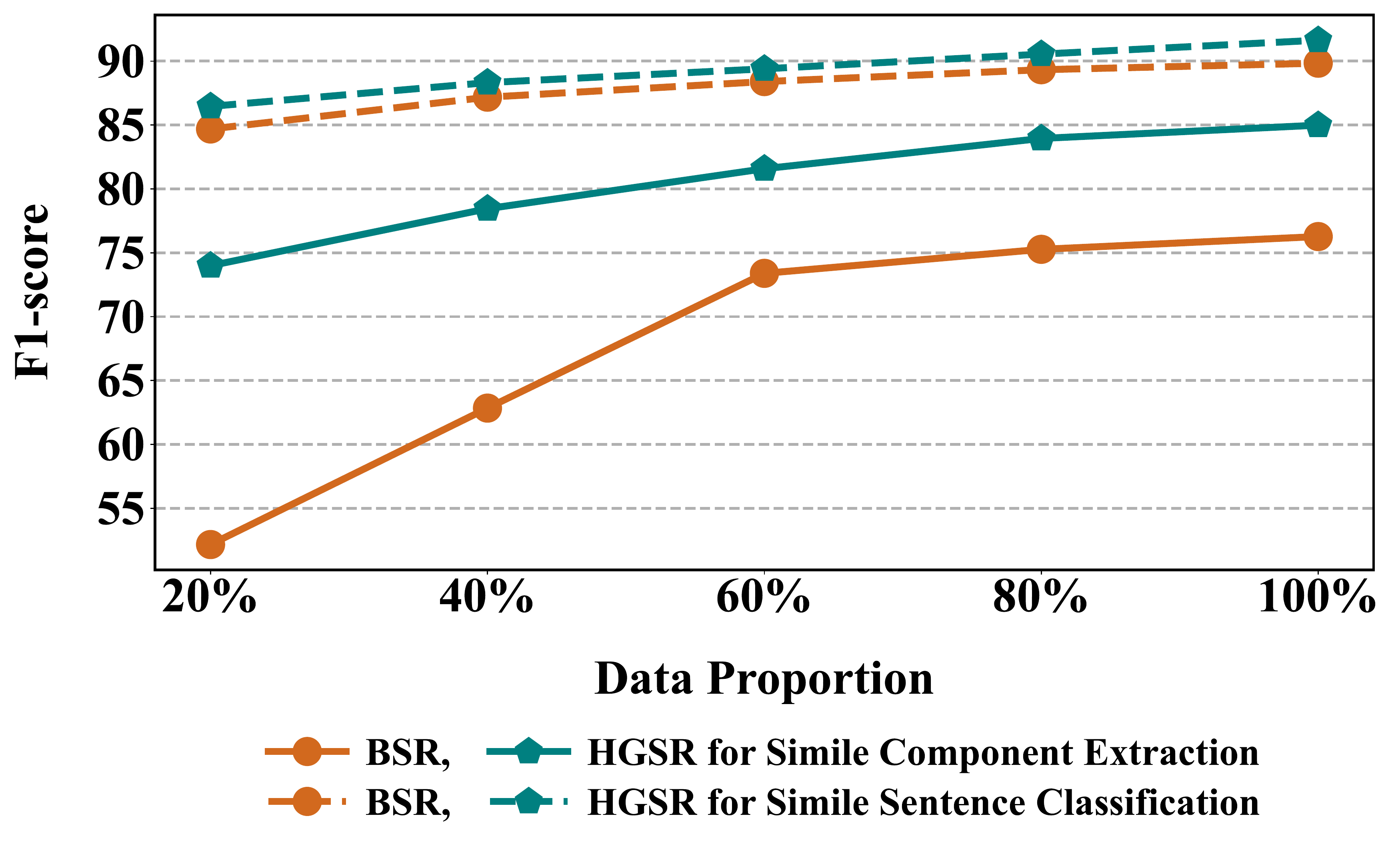}
 \caption{F1 scores on low-resource settings where only a certain percent of data is available for training. Different line colors and styles represent different tasks and systems.}
 \label{data_pro}
\end{figure}

\begin{figure}[t]
 \includegraphics[width=\linewidth]{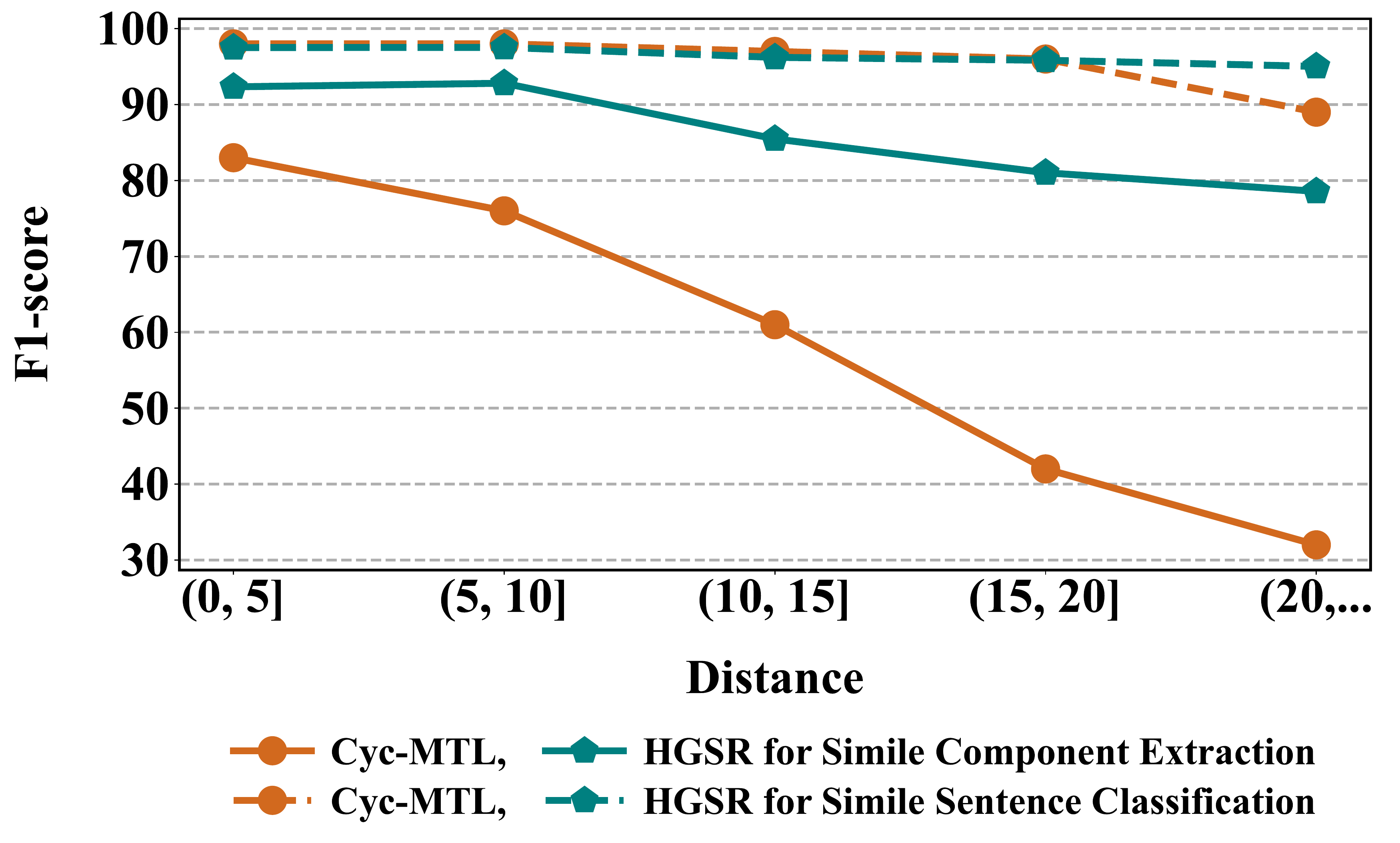}
 \caption{F1 scores on different groups of test instances according to the distance between a tenor and a vehicle.}
 \label{distance}
\end{figure}

\subparagraph{F1 Score against Sentence Length.}
As shown in Figure \ref{length}, we compare our model with \emph{Cyc-MTL} \cite{zeng2020neural} regarding different ranges of sentence length, where we use their provided results that correspond to the reported performance.
Results show that our model is consistency better than \emph{Cyc-MTL} in all groups, and our model always performs better with the increase of sentence length.
This verifies the effectiveness of our dependency-based features for helping handle the long-range dependency problem.

\subparagraph{F1 Score against the Distance between a Tenor and a Vehicle.}
As shown in Figure \ref{distance}, we also analyze the results regarding different groups of distances between a tensor and a vehicle.
We can observe that both models yield descent performances on simile sentence classification, while the challenge is still large for simile component extraction.
For \emph{Cyc-MTL}, the performance in terms of F1 score drops to around 40\% when there are more than 15 words in between, and the number further drops to 30\% if there are more than 20 words.
On the other hand, our model can yield F1 scores of more than 80\% and more than 60\% for these situations, respectively. 
This indicates the robustness of our model on the most challenging cases.

Besides, we analysis models
\textbf{F1 Score on Low Resource Settings} to measure their capabilities when data is insufficient.
As shown in Figure \ref{data_pro}, 
we train several \emph{HGSR} and \emph{BSR} models under the supervision of 20\%, 40\%, 60\%, 80\% and 100\% training data and then evaluate them on the test set. We can observe that the \emph{HGSR} consistently surpass \emph{BSR} in all data settings. Besides, it is encouraging to see that \emph{HGSR} only drops 12.9\% on simile component extraction when data is insufficient (20\%), compared with 31.6\% of \emph{BSR}. And only 40\% training data is enough to train a satisfactory \emph{HGSR}, which even surpasses the \emph{BSR} trained by 100\% training data.
This indicates that the proposed \emph{HGSR} is less data hungry with the help of input-side and decoding features.



\begin{table*}[t]
\centering
\begin{tabular}{l|ccc|ccc}
\toprule[1pt]
\multirow{2}{*}{Model} & \multicolumn{3}{c|}{Sentence Classification}               & \multicolumn{3}{c}{Component Extraction}                   \\ \cline{2-7} 
                       & \multicolumn{1}{c}{Precision} & \multicolumn{1}{c}{Recall} & F1 & \multicolumn{1}{c}{Precision} & \multicolumn{1}{c}{Recall} & F1 \\ \hline

BSR & 87.06 & 93.48 & 90.16 & 74.51 & 78.56 & 76.48 \\
Cyc-MTL \cite{zeng2020neural} & 86.46 & \textbf{95.03} & 90.54 & 75.07 & 79.91 & 77.41 \\ \hline
HGSR              &  \textbf{89.04}     &   94.39    &    \textbf{91.93}   & \textbf{81.86}     &  \textbf{88.37}     & \textbf{85.42}   \\ \bottomrule[1pt]
\end{tabular}%
\caption{The model performance with RoBERTa.}
\label{tab:roberta}
\vspace{-0.5em}
\end{table*}

Finally, we replace the pretrained model from Chinese BERT with a Chinese RoBERTa-wwm \cite{DBLP:conf/emnlp/CuiC000H20}, so as to further investigate the generality of our model. We also apply the Chinese RoBERTa-wwm to \emph{Cyc-MTL}, which is our most competitive baseline. Results are shown in Table \ref{tab:roberta}. We can observe that \emph{HGSR} still surpasses both \emph{Cyc-MTL} and \emph{BSR} on for both sentence classification and component extraction tasks. This verifies that our model is effective with various pretrained models.

\begin{table*}[t]
\centering
\begin{tabular}{llcccc}
\toprule[1pt]
\multicolumn{2}{l}{\multirow{2}{*}{Model}}                          & \multicolumn{1}{c}{\multirow{2}{*}{Sentence Classification (F1)}} & \multicolumn{3}{c}{Component Extraction} \\ \cline{4-6} 
\multicolumn{2}{l}{}                                                & \multicolumn{1}{c}{}                                              & Tenor (F1) & Vehicle (F1) & Overall (F1) \\ \hline
\multicolumn{2}{l}{HGSR}                                             & \textbf{91.64}                                                              & \textbf{90.54}      & \textbf{91.11}        & \textbf{84.99}        \\ \hline
\multicolumn{1}{l|}{\multirow{4}{*}{\ding{172}}} & w/o dependency           & 91.36                                                              & 85.04      & 90.77        & 83.43        \\
\multicolumn{1}{l|}{}                     & w/o POS                  & 91.20                                                              & 87.41      & 89.13        & 83.69        \\ 
\multicolumn{1}{l|}{}                     & w/o definitions          & 90.83                                                              & 85.75      & 89.54        & 82.07        \\ 
\multicolumn{1}{l|}{}                     & w/o subsentence nodes    & 90.64                                                              & 87.62      & 90.91        & 83.83        \\ \hline
\multicolumn{1}{l|}{\multirow{3}{*}{\ding{173}}} & w/o model$_t$             & 91.23                                                              & 89.95      & 86.61        & 83.26        \\
\multicolumn{1}{l|}{}                     & w/o model$_v$             & 91.42                                                              & 85.40      & 90.95        & 83.52        \\ 
\multicolumn{1}{l|}{}                     & w/o model$_p$             & 91.44                                                              & 88.84      & 90.30        & 83.12        \\ 
\hline
\multicolumn{2}{l}{BSR}                                              & 89.84                                                              & 84.13      & 87.18        & 75.12        \\ \bottomrule[1pt]
\end{tabular}
\caption{Ablation study on the main test set, where ``\ding{172}'' and ``\ding{173}'' represent the input features and decoding features, respectively.}
\label{tab:ablation}
\end{table*}

\subsection{Ablation study}
\label{sec:ablation}
To investigate the influence of different features on the model effects, we conducted an ablation study regarding the two major features.

\subparagraph{Input-side features.}
We explore the following variants to investigate the impacts of input-side features:
1) \emph{\textbf{w/o dependency}.} In this variant, we let each word node to connect all other word nodes rather than following the dependency arcs. This is for pinpointing the effect of using dependency information.
2) \emph{\textbf{w/o POS}.} This baseline does not utilize the POS information, where each subsentence node connects all word nodes rather than only noun nodes.
3) \emph{\textbf{w/o definitions}.} 
In this variant, only the input sentence will be fed into the model.
4) \emph{\textbf{w/o subsentence nodes}.}~~In this variant, we merge the two subsentence nodes into one global node and the initial state of the global node is the representation of ``[CLS]''.

As shown in group \ding{172} of Table \ref{tab:ablation}, consistent performance decrease on both subtasks is witnessed after removing each input-side features. 
These results verify the effectiveness of using dependency tree, POS information, word definitions as well as the subsentence nodes. 
Among these factors, we observe that the subsentence nodes cause the greatest impact on sentence classification, which indicates the effectiveness of highlighting the difference between the two sides of the comparator.
Besides, the word definitions give the greatest impact on component extraction. This quite fits our expectation and is consistent with previous observations \cite{niculae2014brighter}.



\subparagraph{Decoding features.}
We further study the effectiveness of the decoding features by removing model$_t$/model$_v$/model$_p$, respectively.

As shown in group \ding{173} of Table \ref{tab:ablation}, we find consistently decline of the model performance when one or more sub models are removed, suggesting that each model learns advantages from other models.
Most importantly, we observe a large decrease (4.5/5.1 F1 points) on Vehicle F1/Tenor F1 when model$_t$/model$_v$ is removed.
These results strongly suggest the importance of utilizing the proposed decoding features.
Besides, we also observe a performance decrease by removing model$_p$, suggesting that it also helps improve the whole system.
By combining both comparison results, we can reach the conclusion: knowledge distillation with the ensemble of all three sub models (model$_t$, model$_v$ and model$_p$) is important for the overall performance of our model.




\vspace{-0.6em}


\begin{table*}[t]
\small
\centering
\begin{CJK}{UTF8}{gbsn}
\begin{tabularx}{\textwidth}{p{1.8cm}<{\centering}p{6cm}ll}
\toprule
Type \& Cnt  &
  Example &
  Cyc-MTL &
  HGSR \\ \midrule
\begin{tabular}[c]{@{}p{1.8cm}<{\centering}@{}}Component dropping\\ (137/77)\end{tabular} &
  \textit{\begin{tabular}[c]{@{}p{6.3cm}@{}}一篇篇\textcolor{RoyalBlue}{作文}、\textcolor{RoyalBlue}{日记}像\textcolor{PineGreen}{泉水}一样从笔下涌出。\\ Pieces of \textcolor{RoyalBlue}{essays} and \textcolor{RoyalBlue}{diaries} gush out from the writing like \textcolor{PineGreen}{spring}.\end{tabular}} &
  \begin{tabular}[c]{@{}l@{}}Type: simile\\ \textcolor{RoyalBlue}{Tenor: \emph{日记(diaries)}}\\ \textcolor{PineGreen}{Vehicle: \emph{泉水(spring)}}\end{tabular} &
  \begin{tabular}[c]{@{}l@{}}Type: simile\\ \textcolor{RoyalBlue}{Tenor: \emph{日记(diaries)}}\\ \textcolor{RoyalBlue}{Tenor: \emph{作文(essays)}}\\ \textcolor{PineGreen}{Vehicle: \emph{泉水(spring)}}\end{tabular} \\ \midrule
\begin{tabular}[c]{@{}p{1.8cm}<{\centering}@{}}Locating error\\ (86/57)\end{tabular} &
  \textit{\begin{tabular}[c]{@{}p{6cm}@{}}乌龟的\textcolor{RoyalBlue}{壳}像\textcolor{PineGreen}{小山}。\\ The \textcolor{RoyalBlue}{shell} of the tortoise is like a \textcolor{PineGreen}{hill}.\end{tabular}} &
  \begin{tabular}[c]{@{}l@{}}Type: simile\\ \textcolor{RoyalBlue}{Tenor: \emph{乌龟(tortoise)}}\\ \textcolor{PineGreen}{Vehicle: \emph{小山(hill)}}\end{tabular} &
  \begin{tabular}[c]{@{}l@{}}Type: simile\\ \textcolor{RoyalBlue}{Tenor:} \textcolor{RoyalBlue}{\emph{壳(shell)}}\\ \textcolor{PineGreen}{Vehicle: \emph{小山(hill)}}\end{tabular} \\ \midrule
\begin{tabular}[c]{@{}p{2cm}<{\centering}@{}}Simile classification error\\ (197/156)\end{tabular} &
  \textit{\begin{tabular}[c]{@{}p{6.3cm}@{}}如果不保护动物，大熊猫迟早会像恐龙一样灭绝。\\ If animals are not protected, pandas will become extinct like dinosaurs sooner or later.\end{tabular}} &
  \begin{tabular}[c]{@{}l@{}}Type: simile\\ Tenor: \emph{熊猫(pandas)}\\ Vehicle: \emph{恐龙(dinosaurs)}\end{tabular} &
  Type: literal \\ \midrule
\begin{tabular}[c]{@{}p{1.8cm}<{\centering}@{}}Redundant component extraction\\ (15/18)\end{tabular} &
  \textit{\begin{tabular}[c]{@{}p{6.3cm}@{}}\textcolor{RoyalBlue}{天空}中没有星星，它像一个巨大的\textcolor{PineGreen}{黑洞}。\\ There are no stars in the \textcolor{RoyalBlue}{sky}, it likes a huge \textcolor{PineGreen}{black hole}.\end{tabular}} &
  \begin{tabular}[c]{@{}l@{}}Type: simile\\ \textcolor{RoyalBlue}{Tenor: \emph{它(it)}}\\ \textcolor{PineGreen}{Vehicle: \emph{星星(stars)}}\\ \end{tabular} &
  \begin{tabular}[c]{@{}l@{}}Type: simile\\ \textcolor{RoyalBlue}{Tenor: \emph{天空(sky)}}\\\textcolor{RoyalBlue}{Tenor: \emph{它(it)}}\\ \textcolor{PineGreen}{Vehicle:\emph{黑洞(black hole)}}\end{tabular} \\ \bottomrule
\end{tabularx}
\end{CJK}
\caption{Case Study on four major types of errors. The gold answers are highlighted with \textcolor{RoyalBlue}{blue} and \textcolor{PineGreen}{green} colors in the ``\emph{Example}'' column. The counts separated by ``/'' in the first column represent the number of mistakes made by \emph{Cyc-MTL} and \emph{HGSR} for each error type, respectively.}
\label{case}
\vspace{-1.0em}
\end{table*}
\subsection{Case Study}
Based on the ground-truth results, we analyze the prediction results of \emph{Cyc-MTL} and \emph{HGSR} on the test set, then we group the errors into four major types and count their respective occurrences. The four types of errors are: \emph{component dropping}, where an output misses important simile components; \emph{locating error}, where a wrong span is extracted as a simile component; \emph{simile classification error}, where a simile/literal sentence is erroneously considered as the other type; \emph{redundant component extraction}, where extra spans (in addition to the gold spans) are extracted as simile components. For better illustration, we list several representative cases, as shown in Table \ref{case}. 
\par In general, \emph{HGSR} is much better than \emph{Cyc-MTL} regarding the first 3 types of errors. It can particularly reduce the \emph{locating error} issue, where the error reduction is more than 50\%. Besides, it also largely alleviates the \emph{component dropping} issue. Both situations are highly correlated with the dependency information, which can be well represented by \emph{HGSR}. One typical example is the second case in Table \ref{case}. \emph{Cyc-MTL} extracts ``tortoise'', probably because it is the main entity in the sentence. On the other hand, \emph{HGSR} correctly extracts ``shell'', which directly connects with the comparator ``like'' in the dependency tree. For the last case, \emph{HGSR} predicts both ``sky'' and  ``it'', where they form a coreference relation.
We consider the prediction of \emph{HGSR} as reasonable, though the reference does not contain ``it''.

\section{Related work}
Our related work mainly includes the studies of simile recognition and heterogeneous neural network for NLP.

\subparagraph{Simile Recognition.}
Early studies mainly focus on classifiers based on manually created patterns and syntactic features. For example, \citet{2008Computation}
adopts a maximum entropy model to recognize simile sentences. In addition, \citet{niculae2013comparison} uses syntactic patterns to extract potential simile components. 
\citet{niculae2014brighter} aims to distinguish between figurative and literal by using a series of linguistics cues as features. However, such pattern-based methods can not deal with the sentences with complex syntactic structures. Inspired by promising results of deep neural networks, \citet{liu2018neural} and \citet{simile2019} introduce multitask learning into simile recognition. Furthermore, \citet{zeng2020neural} proposes Cyc-MTL that considers the inter-correlation between different subtasks of simile recognition.

\subparagraph{HGNN for NLP.}Recently, heterogeneous graph neural network (HGNN) has been shown effective in several NLP tasks, such as relation extraction\cite{DBLP:conf/emnlp/Zhang0M18},
sentence ordering\cite{DBLP:conf/ijcai/YinSSZZL19, lai2021improving, DBLP:journals/jair/YinLSZHYS21}, graph node classification  \cite{wang2019heterogeneous}, question answering \cite{tu2019multi}, intent recommendation \cite{fan2019metapath}, text classification \cite{wang2021cross}, event detection \cite{wang2018cross,cui-etal-2020-edge}, machine translation \cite{DBLP:conf/acl/YinMSZYZL20} and document summarization \cite{wang2020-heterogeneous}.
To our knowledge, this is the first attempt to explore HGNN for simile recognition. Besides,
different from previous work that mainly focuses on encoding one type of features (e.g. dependency tree), ours explores more relevant features to enhance graph representations.

\section{Conclusion}
In this paper, we propose \emph{HGSR}, which gets the most out of task features to alleviate the data hunger issue for simile recognition. Concretely, we explore the input-side features and the decoding features. The input-side features, which includes POS tags, dependency tree and word definitions are encoded via heterogeneous graph encoding. For the decoding features, we build two models sequentially extracted simile components in the opposite orders, then force these two models and the basic model which extracts components in parallel to mimic the behavior of their ensemble. During inference time, only one of these models will be used. Experimental results and in-depth analyses demonstrate the superiority of our model under both sufficient and insufficient data settings.

\section*{Limitations}

The limitations of this work are the following aspects:
1) In this work, experiments are conducted only on Chinese due to the availability of descent-scaled annotated data. We will evaluate the proposed model on other languages once the corresponding large-scale datasets are available.
2) Same as previous work \cite{liu2018neural, zeng2020neural}, we only focus on the result of simile recognition itself, ignoring further discussions on its contribution for other tasks. 
3) The proposed HGSR uses model ensemble as the teacher model for knowledge distillation during training, making the training phase not eco-friendly. We plan to explore the lightweight models and investigate alternative eco-friendly plans in the future.

\section*{Acknowledgements}
We would like to thank the anonymous reviewers for their insightful comments and suggestions. This research is supported by National Natural Science Foundation of China (No. 62276219), Natural Science Foundation of Fujian Province of China (No. 2020J06001), and Youth Innovation Fund of Xiamen (No. 3502Z20206059).

\bibliography{custom,anthology}
\bibliographystyle{acl_natbib}




\end{document}